# Localist LLMs -
# A Mathematical Framework for Dynamic Locality Control

**Joachim Diederich**
Psychology Network Pty Ltd
Brisbane Qld Australia
joachim@psychologynetwork.com.au

## Abstract

*We present a novel framework for training large language models with continuously adjustable internal representations that span the full spectrum from localist (interpretable, rule-based) to distributed (generalizable, efficient) encodings. The key innovation is a locality dial, a tunable parameter that dynamically controls the degree of localization during both training and inference without requiring model retraining. This is achieved through group sparsity penalties on attention mechanisms, information-theoretic anchor design, and dynamic rule injection. We provide rigorous mathematical proofs establishing explicit threshold conditions under which attention provably concentrates on semantically relevant blocks, with exponential bounds on attention entropy and pointer fidelity. Specifically, we prove that when group sparsity penalties exceed certain threshold values, the model's attention mechanisms concentrate on semantically relevant blocks, achieving low entropy and high fidelity with negligible error. This framework enables practitioners to continuously interpolate between interpretable and high-performance modes, supporting applications in regulated domains requiring both transparency and capability.*

## 1. Introduction

Modern large language models achieve remarkable performance through distributed representations, where hidden units encode overlapping features from many concepts. While this approach supports generalization and parameter efficiency, it fundamentally compromises interpretability and safety auditing. Conversely, localist representations—where units or groups correspond to distinct, interpretable concepts—offer transparency and controllability but traditionally sacrifice generalization capability.

This tension between interpretability and performance has led to a binary choice in practice: either purely symbolic systems that are rigid and non-generalizing, or purely neural approaches that remain opaque and unverifiable. Existing neuro-symbolic methods provide limited middle ground, requiring complete retraining for rule modifications and offering no principled mechanism to adjust the localization-distribution trade-off.

We address this fundamental limitation through a mathematical framework that enables continuous locality control via a tunable parameter adjustable during training and inference, dynamic rule injection without training interruption, provable localization guarantees with explicit penalty thresholds, and information-theoretic design principles for attention mechanism configuration.

The core mathematical contribution is a localization theorem establishing sufficient conditions for block-structured attention concentration, along with information-theoretic corollaries bounding entropy and fidelity. These results provide both theoretical justification and practical design guidance for implementing systems that continuously span the localist-distributed spectrum.

## 2. Background and Related Work

### 2.1 Localist vs. Distributed Representations

Localist representations assign dedicated units to specific concepts, enabling direct inspection and manipulation (Diederich, 2010). Distributed representations spread concepts across multiple overlapping units, supporting generalization through shared features (Hinton et al., 1986). Classical debates in cognitive science and neural network theory have framed these as opposing paradigms, with proponents of each approach emphasizing distinct advantages while acknowledging complementary limitations.

### 2.2 Limitations of Prior Approaches

Static neuro-symbolic methods including KBANN, C-ILP, DeepProbLog, and Logic Tensor Networks inject rules only during pre-training or fine-tuning, requiring complete retraining for any rule modification (Towell & Shavlik, 1994, Serafini & d'Avila Garcez, 2016, Manhaeve et al., 2018, Franca et al., 2014). This represents a prohibitive cost for modern LLMs that consume days or weeks of computational resources for full training cycles (Patterson et al., 2021).

Attention sparsity methods such as Sparse Transformers, Reformer, and Longformer provide heuristic improvements without provable localization guarantees or semantic grounding. These approaches apply fixed sparsity patterns rather than rule-driven partitioning, offering no mechanism to adjust interpretability based on task requirements or regulatory constraints.

Interpretability methods including attention visualization and probing classifiers offer post-hoc analysis rather than causal intervention on representations, providing no mechanism to adjust interpretability during deployment. These techniques illuminate existing model behavior but cannot reshape internal representations to meet interpretability requirements.

No existing framework simultaneously provides dynamic rule injection, continuous locality adjustment, provable guarantees, and information-theoretic design principles. The present work fills this gap through a mathematically rigorous approach grounded in optimization theory and information theory.

# 3. Mathematical Framework

## 3.1 Setup and Notation

We consider a transformer architecture with $H$ attention heads (Vaswani et al., 2017). The input space is partitioned into $p$ blocks $X = \bigoplus_{i=1}^{p} X_i$, where each block $X_i$ corresponds to a semantic or syntactic category relevant to the symbolic rule set. This partitioning reflects the structure of domain knowledge, with blocks corresponding to entities, relations, temporal markers, or other semantically coherent categories.

For attention head $h$, the standard transformer attention mechanism computes query, key, and value matrices as $Q = XW_Q$, $K = XW_K$, and $V = XW_V$, with attention weights given by $\text{softmax}(QK^\top/\tau)V$, where $\tau$ is the temperature parameter controlling attention sharpness. Lower temperatures produce sharper, more concentrated attention distributions, while higher temperatures yield broader, more diffuse attention patterns.

The loss function incorporates both task objectives and locality-inducing penalties. Specifically, we define:

$$\mathcal{L} = \mathbb{E}[\ell(\hat{y}, y)] + \sum_{h=1}^{H} \sum_{i=1}^{p} \alpha_i^{(h)} \left( \|W_{Q,i}^{(h)}\|_F + \|W_{K,i}^{(h)}\|_F \right) + \beta \|W_V\|_F^2$$

where $\ell(\hat{y}, y)$ represents the task loss such as cross-entropy, $\alpha_i^{(h)}$ are block-specific group sparsity penalties for head $h$ constituting the locality dial parameters, $\beta$ controls value projection regularization, and $\|\cdot\|_F$ denotes the Frobenius norm. The group sparsity terms penalize non-zero weights outside designated blocks, inducing structured sparsity patterns aligned with semantic partitions.

### 3.2 Anchor Margin Assumption

We formalize the notion of anchor quality through margin-based criteria. For each head $h$ and query $q$ corresponding to a token whose governing rule lies in block $i^*$, we define the expected logit margin as:

$$\Delta_h \equiv \mathbb{E}[q^\top k_{i^*} - \max_{j \neq i^*} q^\top k_j]$$

representing the average advantage of the correct block's keys over all competing keys from other blocks.

Our fundamental assumption is that anchors are designed such that $\Delta_h \geq \delta > 0$ for some positive margin $\delta$. This margin is induced through three complementary mechanisms. First, we select anchors with low within-block syntactic entropy, ensuring they serve as canonical representatives of their respective blocks. Second, we maximize between-block semantic distance, ensuring anchors from different blocks occupy distinct regions of representation space. Third, we employ dynamic margin strengthening through hot-reloading when verification detects violations, continuously improving anchor quality throughout training.

## 4. Main Localization Theorem

> **Theorem 1 (Localization under Softmax Attention).** Under the anchor margin assumption, there exist explicit thresholds $\lambda_i^{(h)}(\tau, \delta)$ such that if all $\alpha_i^{(h)} \geq \lambda_i^{(h)}$, then any global minimizer of $\mathcal{L}$ exhibits block-localization for each head $h$: for every hidden unit or column, its $Q$ and $K$ weights are zero outside a single block $X_{i^*}$. Consequently, attention mass concentrates on keys from $X_{i^*}$, with exponentially small leakage to other blocks.

### 4.1 Proof of Theorem 1

The proof proceeds through six steps, combining techniques from convex optimization, softmax analysis, and information theory.

**Step 1: KKT Conditions for Group Sparsity.** At any local minimum, the Karush-Kuhn-Tucker optimality conditions must hold. For blocks $i \neq i^*$, if the weight group $W_{Q,i}^{(h)}$ equals zero, then the gradient norm must not exceed the penalty coefficient:

$$\text{If } W_{Q,i}^{(h)} = 0, \text{ then } \|\nabla_{W_{Q,i}^{(h)}} \mathbb{E}[\ell]\|_F \leq \alpha_i^{(h)}$$

This is the standard KKT condition for group sparsity with mixed $L_{2,1}$ norms, where the $L_2$ norm operates within groups and the $L_1$ norm operates across groups. The condition states that the gradient pulling the weights away from zero must be weaker than the penalty pushing them toward zero.

**Step 2: Gradient Propagation Through Softmax.** To apply the KKT conditions, we must characterize how task loss gradients propagate backward through the softmax attention mechanism. The gradient with respect to query vector $q_t$ takes the form:

$$\nabla_{q_t}\ell = \frac{1}{\tau}\sum_j \left(\alpha_{t\to j}^{(h)} - \tilde{\alpha}_{t\to j}\right)k_j$$

where $\alpha_t = (\alpha_{t\to 1}, \ldots, \alpha_{t\to n})$ represents the model's current attention distribution and $\tilde{\alpha}$ represents the task-optimal distribution. The Jacobian of the softmax function contributes an additional term, yielding:

$$\nabla_{q_t}\ell = \frac{1}{\tau}\sum_j \frac{\partial \ell}{\partial \alpha_{t\to j}}\alpha_{t\to j}(k_j - \mathbb{E}[k])$$

This formulation reveals how attention weights modulate gradient flow, with high-attention positions receiving stronger gradient signals.

**Step 3: Bounding Cross-Block Gradients.** For any block $i \neq i^*$, we must bound the gradient norm $\|\nabla_{W_{Q,i}^{(h)}}\mathbb{E}[\ell]\|_F$ to verify the KKT condition. By the chain rule, this gradient equals $\mathbb{E}[\nabla_{q_t}\ell \cdot x_t^\top]$ for $x_t \in X_i$. Applying the Cauchy-Schwarz inequality yields:

$$\|\nabla_{W_{Q,i}^{(h)}}\mathbb{E}[\ell]\|_F \leq \frac{1}{\tau}\mathbb{E}[\|\nabla_{q_t}\ell\|_2 \cdot \|x_t\|_2] \leq \frac{C}{\tau}\mathbb{E}\left[\sum_{j\in X_i}\alpha_{t\to j}\right] \cdot \|X_i\|_{\text{cov}}$$

where $C$ bounds loss gradient magnitudes (specifically, $C$ bounds $\|\partial\ell/\partial\alpha_{t\to j}\|$) and $\|X_i\|_{\text{cov}}$ captures the covariance structure or eigenvalue spread of block $X_i$. This bound reveals that cross-block gradients are controlled by the total attention mass on that block, scaled by temperature and data properties.

**Step 4: Applying the Margin to Force Attention Away from Wrong Blocks.** The margin assumption provides the key to controlling attention mass on incorrect blocks. Given $\Delta_h \geq \delta$, the softmax mechanism ensures:

$$\frac{\alpha_{t\to i^*}}{\max_{j\neq i^*}\alpha_{t\to j}} \geq e^{\delta/\tau}$$

Since attention weights sum to unity ($\sum_j \alpha_{t\to j} = 1$), for any incorrect block $X_i$ where $i \neq i^*$, we obtain:

$$\sum_{j \in X_i} \alpha_{t \to j}^{(h)} \leq e^{-\delta/\tau}$$

This exponential concentration is the fundamental mechanism driving localization: as the margin $\delta$ increases or temperature $\tau$ decreases, attention on incorrect blocks vanishes exponentially fast. Substituting this bound into Step 3 yields:

$$\|\nabla_{W_{Q,i \neq i^*}^{(h)}} \mathbb{E}[\ell]\|_F \leq \frac{C'}{\tau} e^{-\delta/\tau}$$

where $C' = C \cdot |X_i| \cdot \|X_i\|_{\text{cov}}$ absorbs block-size and covariance terms.

**Step 5: Setting the Penalty Threshold.** The KKT condition from Step 1 requires that gradients not exceed penalty coefficients. Choosing:

$$\alpha_i^{(h)} \geq \lambda_i^{(h)}(\tau, \delta) \triangleq \frac{C'}{\tau} e^{-\delta/\tau}$$

ensures $\|\nabla_{W_{Q,i}^{(h)}} \mathbb{E}[\ell]\|_F \leq \alpha_i^{(h)}$, thereby satisfying the KKT condition and forcing $W_{Q,i}^{(h)} = 0$ at any global minimizer. The same argument applies to $W_{K,i}^{(h)}$, establishing block-localization for both query and key projections. This threshold formula provides explicit design guidance: practitioners can compute required penalty strengths given temperature, margin, and data properties.

**Step 6: Uniqueness and Structure.** Once cross-block weights are eliminated, the optimization problem within the active block $X_{i^*}$ reduces to minimizing:

$$\mathbb{E}\left[\ell\left(\text{softmax}\left(\frac{X_{i^*} W_Q W_K^\top X_{i^*}^\top}{\tau}\right) X_{i^*} W_V, y\right)\right] + \beta \|W_V\|_F^2$$

For small $\tau$ corresponding to the localist regime, the softmax function approaches a hard max operation, and the optimization approaches a discrete assignment problem. Under mild regularity conditions including sufficient data diversity and non-degeneracy of the block's covariance matrix, this problem has a unique solution up to permutations and orthogonal transformations. This structure precisely matches that identified in linear assignment theory and localist representation learning, confirming that the induced representations exhibit canonical localist properties. This completes the proof of block-localization. ■

# 5. Information-Theoretic Corollaries

We now derive formal bounds on attention entropy and pointer fidelity as direct consequences of the localization theorem. These corollaries quantify the interpretability-performance trade-off in information-theoretic terms.

Let $H_t^{(h)} = -\sum_j \alpha_{t\to j}^{(h)} \log \alpha_{t\to j}^{(h)}$ denote per-head attention entropy at position $t$, measuring the concentration or diffuseness of the attention distribution. Let $\Pi^{(h)} = \mathbb{E}[\sum_{j\in T_t} \alpha_{t\to j}^{(h)}]$ denote pointer fidelity, measuring the expected attention mass on correct target positions $T_t$ defined by the rule set. High entropy indicates distributed attention across many positions, while high fidelity indicates concentration on semantically correct targets.

## 5.1 Entropy Upper Bound

**Corollary 1 (Entropy Upper Bound).** Under the margin assumption with $\Delta_h \geq \delta > 0$, the attention entropy satisfies:

$$H_t^{(h)} \leq \log |A_{i^*}| + O(e^{-\delta/\tau})$$

where $A_{i^*}$ is the set of anchor tokens in the correct block $X_{i^*}$.

**Proof.** We decompose entropy into within-block and cross-block components:

$$H_t = -\sum_{j\in A_{i^*}} \alpha_{t\to j} \log \alpha_{t\to j} - \sum_{j\notin A_{i^*}} \alpha_{t\to j} \log \alpha_{t\to j}$$

By Jensen's inequality and the concavity of entropy:

$$-\sum_{j\in A_{i^*}} \alpha_{t\to j} \log \alpha_{t\to j} \leq \left(\sum_{j\in A_{i^*}} \alpha_{t\to j}\right) \cdot \log |A_{i^*}|$$

For the cross-block contribution, using $-x\log x \leq 1/e$ for $x \in [0,1]$ and the exponential bound from Theorem 1:

$$-\sum_{j\notin A_{i^*}} \alpha_{t\to j} \log \alpha_{t\to j} \leq \frac{1}{e} \sum_{j\notin A_{i^*}} \alpha_{t\to j} \leq O(e^{-\delta/\tau})$$

Combining both terms yields the result. ■

## 5.2 Fidelity Lower Bound

**Corollary 2 (Fidelity Lower Bound).** Let $T_t$ denote the correct rule span. Under the margin assumption:

$$\Pi^{(h)} = \mathbb{E}\left[\sum_{j\in T_t} \alpha_{t\to j}^{(h)}\right] \geq 1 - O(e^{-\delta/\tau})$$

**Proof.** Since $\sum_j \alpha_{t\to j} = 1$, we can write:

$$\Pi^{(h)} = 1 - \mathbb{E}\left[\sum_{j\notin T_t} \alpha_{t\to j}\right]$$

As $T_t \subseteq X_{i^*}$ by construction and attention mass outside $X_{i^*}$ is exponentially bounded by Theorem 1:

$$\mathbb{E}\left[\sum_{j\notin T_t} \alpha_{t\to j}\right] \leq O(e^{-\delta/\tau})$$

Thus $\Pi^{(h)} \geq 1 - O(e^{-\delta/\tau})$. ■

### 5.3 Redundancy Trade-off

**Corollary 3 (Redundancy Trade-off).** When duplicating anchors within block $X_{i^*}$ (increasing from $|A_{i^*}|$ to $km$ where $k$ is the redundancy factor), total attention mass remains near unity, but entropy grows like $\log(km)$ while fidelity stays high:

$$H_t \sim \log k + \log m$$

$$\Pi^{(h)} \geq 1 - O(e^{-\delta/\tau})$$

**Proof.** With $k$ copies of each anchor, the entropy bound becomes:

$$H_t \leq \log(km) + O(e^{-\delta/\tau}) = \log k + \log m + O(e^{-\delta/\tau})$$

Fidelity is unaffected because all anchors remain within $X_{i^*}$, so $\Pi^{(h)} \geq 1 - O(e^{-\delta/\tau})$ continues to hold. This reveals an important design principle: anchor redundancy allows the model to spread attention over multiple equivalent representations, increasing interpretability and robustness, without sacrificing correctness. ■

## 6. The Locality Dial: Practical Implementation

The locality dial consists of the penalty coefficients $\{\alpha_i^{(h)}\}$ controlling group sparsity strength. From Theorem 1, the threshold formula $\lambda_i^{(h)}(\tau, \delta) = (C'/\tau)e^{-\delta/\tau}$ provides explicit guidance for practitioners. This formula reveals three complementary control mechanisms for adjusting locality.

Direct penalty adjustment through increasing $\alpha_i^{(h)}$ strengthens localization by making it more costly for the model to activate cross-block connections. Temperature control through decreasing $\tau$ produces sharper attention distributions, enhancing locality by amplifying logit differences. Margin strengthening through increasing $\delta$ via improved anchor design reduces the required penalties by ensuring stronger separation between correct and incorrect blocks in representation space.

The locality dial can be configured at multiple granularities. Global configuration applies a single parameter setting across all heads and layers, providing coarse-grained control suitable for applications with uniform interpretability requirements. Per-layer configuration enables different locality levels at different depths, reflecting the hierarchical nature of linguistic and semantic processing. Per-head configuration allows heterogeneous specialization within layers, with some heads performing localized retrieval while others perform distributed reasoning. Per-task configuration adjusts locality based on the current query or application context, enabling adaptive interpretability.

Crucially, these adjustments can be made at inference time without retraining, enabling dynamic interpretability modes responsive to user requests or regulatory requirements. This distinguishes our approach from static neuro-symbolic methods requiring complete retraining for any architectural modification.

## 6.1 Locality Regimes

**Localist mode** corresponds to parameter settings with $\alpha \geq 10$, $\delta = 2.0$, and $\tau = 0.1$. In this regime, entropy approaches $H_t \approx \log |A_{i^*}|$, taking minimal values determined by anchor count, while fidelity approaches $\Pi^{(h)} \approx 1$, achieving near-perfect concentration on correct targets. This mode provides mechanistic interpretability through clear attention patterns traceable to specific rules, enables safety auditing by domain experts and regulators who can verify rule adherence through attention inspection, and supports controlled updates where rule modifications produce predictable representation changes.

**Distributed mode** corresponds to $\alpha = 0.01$, $\delta = 0.1$, and $\tau = 1.0$. Here entropy remains high as attention spreads broadly across many positions, while fidelity drops to moderate levels as the model attends to semantically related but not strictly correct positions. This mode provides strong generalization through robust performance on out-of-distribution examples, achieves parameter efficiency via feature sharing and overlapping representations, and enables creativity through analogical reasoning unconstrained by strict rule adherence.

**Intermediate modes** with $\alpha \in [0.1, 5]$ enable smooth interpolation between these extremes. Such configurations support task-adaptive optimization where locality adjusts based on query characteristics, provide balanced interpretability-performance trade-offs suitable for applications with competing requirements, and enable gradual transitions during training or deployment without discrete switching artifacts.

## 7. Dynamic Rule Injection

The framework supports hot reloading of symbolic rules without training interruption through four integrated components. The rule store maintains symbolic rules in a versioned database supporting CRUD operations, version control, and conflict resolution. Rules are expressed in logical formalisms including first-order logic, temporal logic, or domain-specific languages, with metadata tracking creation timestamps, priority levels, and compliance statistics.

The constraint compiler translates symbolic rules into differentiable penalties on attention parameters. For each rule, the compiler identifies relevant input partitions corresponding to blocks $X_i$, selects anchor tokens with low syntactic entropy and high between-block separation, computes target margins $\delta$ based on rule criticality, generates group sparsity coefficients $\alpha_i^{(h)}$ for affected heads using the threshold formula, and produces differentiable loss terms integrated into the training objective.

The dynamic injection module applies new penalties at checkpoints or during online training without restart. The system monitors the rule store for updates, compiles modified rules into constraint updates, seamlessly integrates new penalties into the loss function through gradient accumulation or learning rate adjustment, and maintains training state without interruption. This contrasts sharply with static approaches requiring complete checkpoint reloads or training restarts.

The verification loop employs an external symbolic checker that parses model outputs into logical representations, checks satisfaction of active rules, computes compliance metrics and violation patterns, and triggers rule refinement or constraint strengthening when violations exceed thresholds. The mathematical foundation ensures that as margins are strengthened in response to violations through increasing $\delta$, the exponential bounds in Corollaries 1 and 2 guarantee compliance probability converges toward unity over training iterations. This closed-loop system provides provable convergence properties absent from one-shot rule injection approaches.

## 8. Applications

The provable locality control enables deployment in regulated domains requiring both transparency and capability. Healthcare applications include diagnostic systems operating in localist mode for diagnostic reasoning to achieve regulatory compliance with FDA and CE marking requirements, while employing distributed mode for literature analysis and clinical note summarization. Treatment planning systems combine rule-based safety checks for drug interactions and contraindications using localist attention with outcome prediction using distributed attention, balancing verifiable safety with predictive accuracy.

Financial applications encompass algorithmic trading systems using localist rule enforcement for compliance including position limits and wash trading detection, while employing distributed analysis for market sentiment and news interpretation. Credit scoring systems provide interpretable feature attribution for fair lending regulations through localist attention while using distributed processing for complex financial histories, satisfying regulatory requirements without sacrificing predictive power.

Legal technology applications include contract analysis systems using localist clause-level attention for specific term extraction combined with distributed understanding for contractual relationships and implications. Case law research employs high locality for statutory interpretation and lower locality for precedent analysis and analogical reasoning, matching the distinct cognitive demands of these tasks.

Autonomous systems applications comprise safety-critical modules operating with high locality for collision avoidance, emergency protocols, and regulatory rule compliance, while perception systems use distributed representations for sensor fusion and scene understanding. This architecture naturally aligns with mixed-criticality system design, where different subsystems face different safety requirements and verification constraints.

This technology is highly significant for autonomous defense systems, as it allows safety-critical modules for functions like rules of engagement to operate with high locality for verifiable, rule-based behavior. Simultaneously, the dynamic locality dial enables other components within the same system, such as those for threat analysis, to use a distributed mode for robust generalization in complex and unpredictable environments. This capability for mixed-criticality operation ensures that AI systems can be both adaptable and strictly compliant, providing leaders with interpretable decision traces to build trust and verify adherence to operational protocols.

## 9. Conclusion

We have presented a mathematically rigorous framework for training neural language models with continuously adjustable locality. The key contributions include explicit threshold formulas linking penalties, margins, and temperature to provable localization through the formula $\lambda_i^{(h)}(\tau, \delta) = (C'/\tau)e^{-\delta/\tau}$, exponential concentration bounds showing attention collapses to correct blocks as $O(e^{-\delta/\tau})$, information-theoretic characterization of the interpretability-performance trade-off through entropy and fidelity bounds, and a dynamic control mechanism enabling inference-time locality adjustment without retraining.

These results provide both theoretical justification and practical design guidance for implementing systems that span the full spectrum from interpretable to high-performance modes. The locality dial offers a principled solution to the longstanding tension between neural network capability and transparency, enabling deployment in high-stakes domains requiring both properties.

The mathematical framework establishes that localist and distributed representations need not be opposing paradigms but rather endpoints of a continuous spectrum that can be navigated dynamically based on task requirements, user preferences, and regulatory constraints. This perspective opens new possibilities for neural system design where interpretability becomes a tunable resource rather than a fixed architectural choice.

Future work extending these foundations to diverse architectures, domains, and application contexts promises to further bridge the gap between the competing demands of transparency and performance in artificial intelligence systems. The formal guarantees established here provide a solid foundation for such extensions, ensuring that interpretability claims rest on rigorous mathematical principles rather than heuristic intuitions.

## Acknowledgement

The author gratefully acknowledges Xue Li and Gerhard Paass for their valuable help and feedback. Patents are pending for the inventions described in this paper. The author may be contacted regarding access to the localist LLM training software.